%% file: sn-article.tex
\documentclass[pdflatex,sn-mathphys-num,iicol]{sn-jnl}

\usepackage{graphicx}%
\usepackage{multirow}%
\usepackage{amsmath,amssymb,amsfonts}%
\usepackage{amsthm}%
\usepackage{mathrsfs}%
\usepackage[title]{appendix}%
\usepackage{textcomp}%
\usepackage{manyfoot}%
\usepackage{booktabs}%
\usepackage{algorithm}%
\usepackage{algorithmicx}%
\usepackage{algpseudocode}%
\usepackage{listings}%

\usepackage{graphicx}
\graphicspath{{figs/}}
\usepackage{multirow}
\usepackage{bbding} 
\usepackage{color}
\usepackage{float}
\usepackage[dvipsnames]{xcolor} 
\definecolor{ForestGreen}{RGB}{34,139,34} 
\usepackage{xspace}
\usepackage[table,xcdraw]{xcolor}
\usepackage{hyperref}
\usepackage{amsmath}
\usepackage{wrapfig}
\usepackage{subcaption}
\usepackage{tabularx}
\usepackage[normalem]{ulem}
\useunder{\uline}{\ul}{}
\newcommand{\ours}{OneClip-RAG\xspace}

\usepackage{etoolbox} 

\makeatletter
\patchcmd{\corrauthor}{\textsuperscript{*}}{\textsuperscript{\dag}}{}{}
\patchcmd{\corrauthorinfo}{\textsuperscript{*}}{\textsuperscript{\dag}}{}{}
\makeatother

\theoremstyle{thmstyleone}%
\theoremstyle{thmstyletwo}%

\theoremstyle{thmstylethree}%

\begin{document}

\title[Article Title]{Towards Effective Long Video Understanding of Multimodal Large Language Models via One-shot Clip Retrieval}


\author[1]{\fnm{Tao} \sur{Chen}}\email{\{taochen, jushaobo, qiong, fangchenxin, kunzhang, pengjun\}@stu.xmu.edu.cn}

\author[1]{\fnm{Shaobo} \sur{Ju}}

\author[1]{\fnm{Qiong} \sur{Wu}}

\author[1]{\fnm{Chenxin} \sur{Fang}}

\author[1]{\fnm{Kun} \sur{Zhang}}

\author[1]{\fnm{Jun} \sur{Peng}}

\author[1]{\fnm{Hui} \sur{Li}}\email{\{hui, rrji\}@xmu.edu.cn}

\author*[1]{\fnm{Yiyi} \sur{Zhou}}\email{zhouyiyi@xmu.edu.cn}

\author[1]{\fnm{Rongrong} \sur{Ji}}

\affil[1]{\orgdiv{Key Laboratory of Multimedia Trusted Perception and Efficient Computing, Ministry of Education of China}, \orgname{Xiamen University}, \orgaddress{\postcode{361005}, \country{P.R. China}}}


\abstract{Due to excessive memory overhead, most \emph{Multimodal Large Language Models} (MLLMs) can only process videos of limited frames. In this paper, we propose an effective and efficient paradigm to remedy this shortcoming, termed \emph{\textbf{One}-shot video-\textbf{Clip} based \textbf{R}etrieval-\textbf{A}ugmented \textbf{G}eneration} (OneClip-RAG). Compared with existing video RAG methods, OneClip-RAG makes full use of the merits of video clips for augmented video understanding in terms of both knowledge integrity and semantic coherence. Besides, it is also equipped with a novel query-guided video chunking algorithm that can unify clip chunking and cross-modal retrieval in one processing step, avoiding redundant computations. To improve instruction following, we further propose a new dataset called \textbf{SynLongVideo} and design a progressive training regime for OneClip-RAG. OneClip-RAG is plugged into three recent MLLMs and validated on a set of long-video benchmarks. Experimental results not only show the obvious performance gains by OneClip-RAG over MLLMs, \emph{e.g.}, boosting Qwen3-VL 8B to the level of GPT-5 on MLVU, but also show its superior efficiency in handling long videos.  
\emph{e.g.}, enabling LLaVA-Video understand up to an hour of videos in less than 1.2 minutes on a single 4090 GPU.
Our code is released at: \href{https://github.com/Tao-Chen-xmu/OneClip-RAG}{\textcolor[rgb]{0,0.2,0.5}{OneClip-RAG}}.}




\maketitle

\input{sec/1_intro}

\input{sec/2_related_work}
\input{sec/3_method}

\input{sec/4_experiment}

\input{sec/5_conclusion}


\bibliography{sn-bibliography}

\end{document}

%% file: sec/1_intro.tex
\section{Introduction}
\label{sec:intro}

In recent years, the great success of \emph{Large Language Models} (LLMs)~\cite{ZhengC00WZL0LXZ23, qwen2, cai2024internlm2, abs-2407-21783} sparks an influx of interest in extending them to multimodal learning, \emph{i.e.}, MLLMs~\cite{Zhu0SLE24}. After culminating in various image-language tasks~\cite{zhou2019plenty,zhou2021trar,luo2024towards,luo2022towards}, recent endeavors turn to the exploration of MLLMs in the domain of video understanding~\cite{ZhangLB23,0002WH00LWX0L0024}. Compared to image-language tasks, the video-based ones are often more challenging, mainly due to the understanding of long and continual video content~\cite{abs-2311-10122}.

\begin{figure*}[t]
  \centering
   \includegraphics[width=\linewidth]{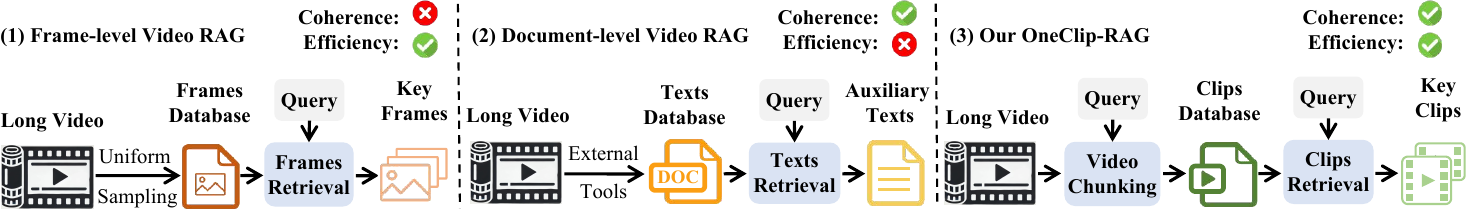}
   \caption{Comparisons between existing video RAG strategies and our OneClip-RAG. OneClip-RAG unifies video chunking and clip retrieval in one unified paradigm based on cross-modal similarities, providing coherent frames for efficient long video understanding.}

   \label{fig:preview}
\end{figure*}

For video-based tasks, most existing MLLMs~\cite{abs-2311-10122,0002WH00LWX0L0024,0001RKK24,abs-2404-16994} are trained and tested with a small number of video frames due to the limitation of GPU memory overhead. Concretely, existing MLLMs~\cite{chen2024far,abs-2409-12191} often use hundreds of visual tokens to represent an image, thereby enhancing their general capability on various image-language tasks at different granularities~\cite{abs-2306-13394,LuBX0LH0CG024}. However, this setting becomes prohibitively expensive for video tasks. 
For instance, to process 16 video frames, InternVL3.5-8B~\cite{abs-2508-18265} needs to consume 15.6 times TFLOPs and 3.4 times GPU memory than its image-question answering. 
In this case, a prevailing but compromise approach for video-MLLMs is to uniformly sample a limited number of video frames, which however is prone to losing key information~\cite{abs-2311-10122,abs-2305-06355,abs-2404-03413,zhang2024llavanext-video,abs-2404-16994}.

The advancement of \emph{retrieval-augmented generation} (RAG) in LLMs~\cite{ShiMYS0LZY24,AsaiWWSH24} yields an effective way to solve this issue. 
In particular, long-sequence modeling is a shared problem for Transformer-based models~\cite{abs-2004-05150,XieCLDD24,abs-2406-16852}, holding similar challenges as discussed above. In terms of LLMs, RAG is often a viable solution to tackle long document understanding in addition to the optimization of the self-attention mechanism~\cite{DaoFERR22,abs-2407-08608,wu2026not}. Via retrieving the most relevant text snippets, an LLM can correctly answer questions based on a limited length of context knowledge, which is often free in long-sequence tuning~\cite{AsaiWWSH24,abs-2401-15884,CuconasuTSFCMTS24}. In this case, it is natural to regard a long video as an external knowledge base and provide the key visual information as the instruction demands, thereby avoiding the excessive computation of all video frames~\cite{ArefeenDUC22,abs-2411-13093}.

Video RAG paradigms have been recently attempted~\cite{ZhangZYML24,abs-2406-12846,abs-2407-15047}, but we find that existing solutions still encounter several challenges. Firstly, most endeavors focus on the key frame selection for MLLMs according to a given instruction~\cite{abs-2407-15047,abs-2406-12846}, as shown in Fig.~\ref{fig:preview}-a. Compared with key frames, we argue that video clips are the better choice as knowledge fragments, which contain more complete and continual semantics, avoiding potential conflicts between selected frames, \emph{i.e.}, content coherence. Secondly, the effective modeling of video clips still needs more exploration. Some recent efforts also retrieve video clips for MLLMs~\cite{abs-2310-19773,ZhangZYML24,AtaallahSASZDZSE24}. But in practice, they often resort to video captioning using additional MLLMs, \emph{e.g.}, MiniGPT4-video~\cite{abs-2404-03413}, so as to reduce the difficulty of cross-modal alignment, \emph{i.e.}, using text retrieval as shown in Fig.~\ref{fig:preview}-b. However, this solution not only requires additional MLLMs and introduces more computational and memory overhead, which also conflicts with the goal of efficient video understanding. Overall, how to effectively and efficiently implement video-clip RAG for MLLMs still remains an open challenge.

In this paper, we propose an effective and efficient method for long video understanding of MLLMs, termed \emph{\textbf{One}-shot video-\textbf{Clip} based \textbf{R}etrieval-\textbf{A}ugmented \textbf{G}eneration} (OneClip-RAG). 
Compared with existing video-RAG methods~\cite{ZhangZYML24,abs-2406-12846,abs-2407-15047}, our approach turns to video clips for the knowledge augmentation of video-MLLMs, providing more coherent knowledge context. 
Besides, we also equip our \ours with an innovative and efficient video chunking algorithm, which conducts query-guided clip chunking for more integrated video content.  Notably, this chunking method can be further combined with popular visual-language (VL) embedding models, \emph{e.g.}, CLIP~\cite{RadfordKHRGASAM21} and SigCLIP~\cite{ZhaiM0B23}, to unify video chunking and clip retrieval in one processing step, avoiding the overuse of additional models and operations.

In addition, we also propose a new dataset to improve the instruction following capability of VL embedding models, termed \emph{SynLongVideo}.
Although the popular VL embedding models like CLIP have been widely used in video-RAG methods~\cite{abs-2411-13093,liu2025bolt}, we notice that they are still inferior in following question-like instructions, especially the ones for video clip retrieval. It is because they are trained on plain image-caption pairs.
Thus, SynLongVideo is used to enhance their capability of cross-modal video clip retrieval. In practice, SynLongVideo exploits the available short video-question pairs to synthesize long video retrieval examples through semantic-based data mixups, \emph{i.e.}, combining short videos to form a synthesized long video based on \emph{visual relevance} and \emph{instruction divergence}.
In this case, these synthesized examples can serve to training the instruction following capability of VL embedding models on long video understanding tasks.
Based on SynLongVideo, we also carefully design a progressive training regime for \ours.

To validate \ours, we apply it to three advanced MLLMs, namely LLaVA-Video~\cite{ZhangWLLMLL25}, Qwen2.5-VL~\cite{Qwen2.5-VL} and Qwen3-VL~\cite{Qwen3-VL}, and conduct extensive experiments on several competitive long VideoQA benchmarks, including LongVideoBench~\cite{abs-2407-15754}, MLVU~\cite{abs-2406-04264}, LVBench~\cite{abs-2406-08035} and Video-MME~\cite{FuDLLRZWZSZCLLZ25}. The experimental results show that, as a plug-and-play component, \ours can greatly improve the long video understanding capability of MLLMs without structure tweaks or dedicated tuning, \emph{e.g.}, helping Qwen3-VL 8B approach the GPT-5~\cite{abs-2601-03267} level on MLVU.
Moreover, \ours enables LLaVA-Video to understand hour-long videos with only 1.2 minutes on average, showing its great efficiency in handling long videos, well supporting its practical applications.

Conclusively, our contribution is threefold:

\begin{itemize}
    \item We study the long video understanding problem of MLLMs from the perspective of video clip, and propose an effective and efficient method called \ours.  
    \item We build a new dataset called \emph{SynLongVideo} with short-video mixups to improve the instruction following capability of VL embedding models, and a coarse-to-fine training regime is also proposed for our \ours.
    \item As a plug-and-play method, \ours can greatly improve the performance of different MLLMs on long video understanding benchmarks, supporting the hour-long video inference using only one 4090 GPU.
\end{itemize}

%% file: sec/2_related_work.tex
\section{Related Work}
\label{sec:rw}

The great success of MLLMs also advances the research of video understanding~\cite{0002WH00LWX0L0024,0001RKK24,11146594,SongCWZZWCG0ZLH24}. Borrowing the principle of image-based MLLMs~\cite{11301404,11086409,luo2024moil}, recent efforts are also devoted to video-MLLMs~\cite{11433110,10948357,10874219} based on the powerful LLMs~\cite{ZhengC00WZL0LXZ23}. In particular, 
VideoChat2~\cite{0002WH00LWX0L0024} introduces a progressive training paradigm with multi-modal instructions, bridging LLM with the visual encoder.
Video-ChatGPT~\cite{0001RKK24} averages frame-level features across temporal and spatial dimensions, respectively, which are then jointly learned with text features in LLMs.
More recent state-of-the-art models like Qwen3-VL~\cite{Qwen3-VL} have achieved remarkable performance improvements by scaling both model parameters and training data. 
Despite these advances, most video-MLLMs often adopt uniform frame sampling to avoid excessive memory overhead~\cite{0002WH00LWX0L0024,0001RKK24}, which however tends to lose key visual semantics for long video understanding. 

To tackle the above challenge, some recent progresses resort to video retrieval, which employ external modules to process large amounts of content before passing it into Video-MLLMs.
In particular, 
BOLT~\cite{liu2025bolt} prioritizes query-relevant frames while preserving selection diversity via inverse transform sampling.
Q-Frame~\cite{abs-2506-22139} allocates varying resolutions to selected keyframes based on their query relevance.
AKS~\cite{abs-2502-21271} adopts a recursive keyframe selection strategy considering both the question-frame relevance and the temporal coverage.
However, as argued above, these video frame selection strategies may face challenges in maintaining temporal and content coherence across selected frames. 
In light of these limitations, some recent works turn to clip-based RAG~\cite{11153868,11397184}. Goldfish~\cite{AtaallahSASZDZSE24} partitions a long video into uniform clips and encodes each clip into text descriptions, subsequently retrieving the top-K relevant descriptions for an LLM. 
MemVid~\cite{abs-2503-09149} proposes a moment retrieval framework using a ``memorizing-reasoning-retrieving-focusing'' pipeline.
One similar work to our \ours is Video-LLaMB~\cite{abs-2409-01071}, which uses scene-based video clip segmentation for external memory retrieval. Compared with it, OneClip focuses on query-related clip chunking and retrieval.
Although these clip-based RAG systems offer great advantages in integrating relevant multimodal information, they also impose substantial computational and time overheads when building the knowledge database, particularly for long videos.

\begin{figure*}[t]
  \centering
   \includegraphics[width=\linewidth]{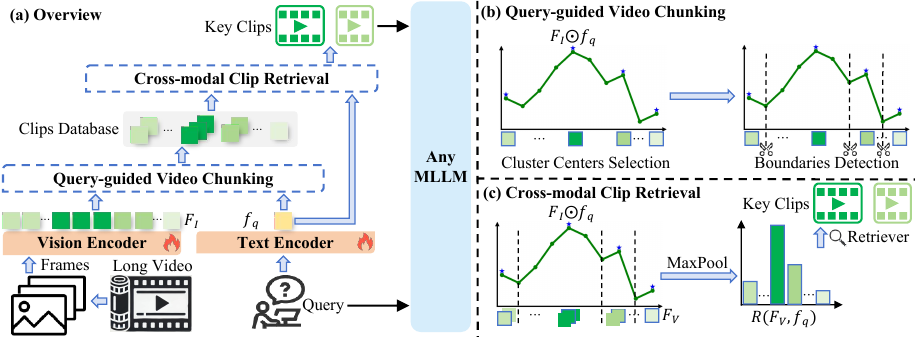}
   \caption{Overview of OneClip-RAG. (a) As a plug-and-play design, OneClip-RAG first performs clip chunking based on the given video and input instruction,  and then selects the most relevant video clips for augmented video understanding of MLLMs. (b) OneClip-RAG uses the cross-modal similarities between frames and text instructions to depict the changes of video content, and then determines the boundaries for video clip chunking. (c) OneClip-RAG can directly select the most relevant clips for MLLMs, requiring no additional models.}
   \label{fig:overview}
\end{figure*}

%% file: sec/3_method.tex
\section{Method}

\subsection{Overview}
\label{sec:overall}

In this paper, we propose a novel approach termed OneClip-RAG for the efficient and effective long video understanding of MLLMs, of which structure is illustrated in Fig.~\ref{fig:overview}.
\ours considers video clips as augmented knowledge for MLLMs, and is also equipped with an innovative query-guided video chunking algorithm that can unify clip chunking and cross-modal retrieval in one procedure.

In particular, given a long video $V$ and a text instruction $Q$, existing video-MLLMs often uniformly sample a few of frames as the input images, denoted as $V'=\{I_1,I_2,...,I_l\}$. And the objective of video-MLLMs is defined by 
\begin{equation}
p(A \mid V', Q) = \prod_{i=1}^L p(a_i \mid V', Q, A_{<i}),
\end{equation}
where $p$ denotes the probabilities of the predicted word, $A=\{a_1,...,a_L\}$ is the answer sequence and $L$ is its length. $A_{<i}$ denotes the answer subsequence before the $i$-th step.

In terms of long video understanding, this compromised solution is prone to losing key information~\cite{liu2025bolt,abs-2502-21271}.
An alternative way is to regard the video as a knowledge base, and retrieve most relevant information for MLLMs~\cite{abs-2407-15047}. Following this idea, the objective of \ours, as a plug-and-play component, is to find out the most relevant clips that can facilitate the correct prediction:
\begin{equation}
    \begin{aligned}
        & \arg\max_{\theta} \, p(A \mid V_{clip},Q), \; \\
        & \text{where} \; V_{clip} = \text{OneClip}(V).
    \end{aligned}
\end{equation}
Here, $\theta$ denotes the parameters of the VL embedding model in OneClip, and $V_{clip}=\{I_k,I_{k+1},...,I_{k+l}\}$ is the query-relevant and continual video frames, \emph{i.e.}, video clips. 

In practice, \ours will act as both a video chunker and a clip retriever for better efficiency, which is supported with a novel query-guided video chunking algorithm.

\subsection{Cross-modal Video Chunking and Retrieval}
\label{sec:cvsr}
One special property of \ours is to 
unify video chunking and instruction-aware clip retrieval into a single process.
The main intuition is that the cross-modal similarities computed by the VL embedding model can not only be used for cross-modal retrieval, but also reflect the query-related semantic changes in video content. In this case, we can exploit this property to avoid redundant computations. 

\noindent\textbf{Query-guided Video Chunking.}  
To achieve the above targets, we propose a novel query-guided chunking algorithm as illustrated in Fig.~\ref{fig:overview}-b.
Compared with existing scene-based chunking methods~\cite{abs-2409-01071}, this query-guided solution can help to capture clips that involve content changes related to the instruction, \emph{e.g.}, the questions may cover multiple scene transitions.

Specifically, given a video $V$, we sample $t$ frames at a shorter interval to densely represent the video, and then compute their cross-modal similarities via the VL embedding model, $S=\{s_1, s_2, ..., s_t\}$. Here, $s_i$ is obtained by
\begin{equation}
s_i = cos (f_{I}^i, f_q),
\end{equation}
where $cos(\cdot)$ denotes the cosine similarity, 
$f_q$ and $f_I^i$ are the query and image features, respectively.

Afterwards, we calculate the peak score~\cite{Hearst97} for each frame $g_i$ and select the frames with highest values as the cluster centers $C$ for the following video chunking:
\begin{equation} \centering 
\begin{aligned} 
& \quad\quad\quad C = \{I_i \mid g_i \in \overset{n}{\underset{l \in 1,...,t}{\arg\max}} \: g_l\}, \\ 
& \text{where} \quad g_i = 2s_i \; 
\begin{aligned}[t] &- \:\min_{j<i}\:\:\{s_j \mid s_j \leq s_i\} \\ 
&- \min_{i<k \leq t}\{s_k \mid s_k \leq s_i\}. 
\end{aligned} 
\end{aligned} 
\label{eq:cluster} 
\end{equation}
Here, $n$ and $t$ are the number of cluster centers and sampled frames, respectively. $j$ and $k$ denote the number of frames to the left and right of $I_i$, respectively. A higher $g_i$ indicates that its near similarities are smaller as shown in Fig.~\ref{fig:overview}-b. 

For each pair of the adjacent cluster centers $(c_j,c_{j+1})$, we aim to identify an optimal boundary point $b^*_j$ that divides the frame sequence between them into two continuous and non-overlapping subsequences. Thus, the problem becomes the alignment of intermediate frames either to the left center $c_j$ or the right center $c_{j+1}$ while preserving continuity. 

We solve this optimization problem using \emph{dynamic programming}~\cite{BerndtC94} to maximize semantic coherence of the two subsequences among all possible boundary candidates:
\begin{multline}
    b^*_j = \underset{b \in \{1, \dots, t_j\}}{\arg\max} \:\: \sum_{k=1}^b (s_k - s_{k+1}) \\
    + \frac{1}{t_j - b} \sum_{l=b+1}^{t_j} (s_{l+1} - s_l),
\end{multline}
where $t_j$ is the number of frame sequence between $c_j$ and $c_{j+1}$. We divide the long video into $n$ semantic clips $\{V_1,...,V_n\}$ according to $n-1$ boundaries $\{b^*_1,...,b^*_{n-1}\}$, which are regarded as the external knowledge base.

Compared with previous scene-oriented chunking methods \cite{MunSHLHLK22,abs-2409-01071}, our query-guided method focuses more on the continual frames relevant to the input query, which are likely to involve multiple scene transitions. Moreover, it can directly adopt the computed VL similarities, and requires no additional comparison between the visual features of all frames~\cite{MunSHLHLK22}, of which computation is quadratic to the VL similarity one. Thus, this property can well facilitate the target of efficient video understanding.

\noindent\textbf{Video clip retrieval.} Given the semantically chunked video clips, we can directly perform cross-modal retrieval.
Specifically, we retrieve $V_{clip}$ according to all clip-instruction relevance scores that has been computed:
\begin{equation}
    \centering
    \begin{aligned}
        r_i &= \max \{s_j \mid I_j \in V_i\}, \\
        V_{clip} &= \{V_i \mid r_i \in \overset{K}{\underset{j=1,...,n}{\arg\max}} \: r_j\},
    \end{aligned}
\end{equation}
where the clip-instruction relevance score $r_i$ of each $V_i$ is obtained by applying max pooling to $S_i=\{s_j \mid I_j \in V_i\}$, and $K$ the number of retrieved clips in $V_{clip}$.
Then its frames are used as the input images for MLLMs, which are much more semantically related to text instruction compared with their default uniform sampling.

\subsection{Coarse-to-fine Instruction Tuning}
\label{sec:ift}

In practice, OneClip-RAG uses popular VL embedding models such as CLIP~\cite{RadfordKHRGASAM21} and SigLIP~\cite{ZhaiM0B23}. However, the instruction following capability of these embedding models on video tasks still has ample room to improve, since they are often trained with only image-caption pairs rather than instruction-video pairs.
Moreover, under conventional RAG settings~\cite{ZhangJBZLLR24,abs-2501-04652}, embedding models tuning is also critical for accurate knowledge retrieval.

In light of this, we introduce a coarse-to-fine training regime to improve the embedding model on instruction-video-clip retrieval.
We first train our OneClip-RAG in a coarse-grained manner using the contrastive learning objective, \emph{i.e.}, directly using the clip frames of other videos as the negative examples.
And the objective of this training can be defined by
\begin{equation}
    \label{eq:cg}
\resizebox{0.89\linewidth}{!}{$
    \mathcal{L}_{cor}=-\frac{1}{m} \sum\limits_{i=1}^{m} \log \frac{\exp(s_i / \tau)}{\exp(s_i / \tau) + \sum_{j \in V_{neg}} \exp(s_j / \tau)}, \\
$}
\end{equation}
where $m$ is the number of frames in the positive clip $V_{pos}$, and $V_{neg}$ denotes the negative clips from other videos. $s_i$ is the question-frame similarity defined above. In principle, Eq.~\ref{eq:cg} is to help the model to  identify the correct video context according to the user queries.

Afterwards, we perform a granular training scheme that requires to target clips with similar contexts.  
Specifically, we only select the negative examples from the other clips $V'_{neg}$ within the same long video:
\begin{equation}
\label{eq:fine_loss}
\resizebox{0.89\linewidth}{!}{$
\mathcal{L}_{fine}=-\frac{1}{m} \sum\limits_{i=1}^{m} \log 
\frac{\exp(s_i / \tau)}{\exp(s_i / \tau) + 
\sum_{k \in V'_{neg}} \exp(s_k / \tau)}.
$}
\end{equation}
The purpose of Eq.~\ref{eq:fine_loss} is to improve the awareness of embedding models in terms of granular text semantics, \emph{i.e.,} identifying the key information among clips with similar context and objects.

During training, these two objectives are conducted sequentially. 
However, implementing the proposed progressive instruction tuning still requires well-labeled instruction-clip pairs. Such data are scarce and labor-intensive, especially for the clip-based RAG of long video understanding.

\begin{figure*}[t]
  \centering
   \includegraphics[width=\linewidth]{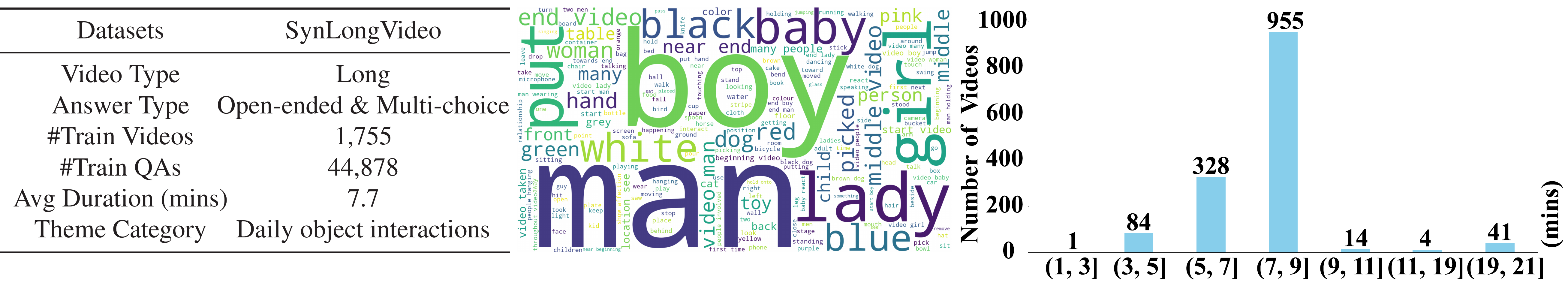}

    \caption{Statistical overview of the proposed SynLongVideo dataset. SynLongVideo aims to improve the instruction-following capability of clip retrieval models for long video understanding. In addition to available long video-question data~\protect\cite{BarmannW22}, it also synthesizes 430 long videos via visually and textually data mix-ups of short videos. The dataset statistics is given in the left table, and its main semantics and data distributions are shown in the middle and right graphs.}
   \label{fig:data_static}

\end{figure*}

\section{Synthesized Long Video Dataset} 
\label{sec:slvd}

To remedy the lack of training data for video-clip RAG, we further propose a new dataset in this paper, termed \emph{Synthesized Long Video Dataset} (SynLongVideo), of which statistics are given in Fig.~\ref{fig:data_static}. In particular, SynLongVideo has 44,878 video-question pairs targeting causal, temporal and descriptive reasoning. It contains 997 long video from QaEgo4D~\cite{BarmannW22} and 430 synthesized ones created based on NeXT-QA~\cite{XiaoSYC21},
which are sourced from everyday interactions under diverse settings such as family activities, social gatherings, and ego-centric scenes. 
The average video length is 7.7 minutes, spanning from 1 to 21 mins.

The construction of SynLongVideo follows two key principles, namely \emph{visual relevance} and \emph{instruction divergence}. The visual relevance measures the similarity among different short videos for the synthesis of long and coherent videos. However, since each video corresponds to multiple instructions, which may share similar or even identical instructions, making the model fail to distinguish relevant clips within the synthesized long videos. To mitigate this issue, we introduce \emph{instruction divergence} to maximize the distinctiveness of instructions across different videos, ensuring that the synthesized long videos contain visually similar semantics but textually distinct instructions. 

\noindent\textbf{Visual relevance.} In this process, we use the visual similarity to retrieve relevant videos and construct a noise candidate set.
Specifically, given the short videos batch, we randomly sample five frames from each video $V_{i}=\{I_i^1,I_i^2,...,I_i^5\}$, and the visual relevance $s(V_i,V_j)$ is obtain by
\begin{equation}
    \begin{aligned}
        s(V_i,V_j) &= cos(f^{v}_i,f^{v}_j), \\
        \text{where} \:\:\: f^{v}_i &= \text{AvgPool}(F_i^v),
    \end{aligned}
\end{equation}
where the video representation $f^{v}_i$ is obtained by average pooling the frame features $F_i^v$ of video $V_i$. For each $V_i$, we retrieve the 16 most similar videos based on $s(V_i,V_*)$ to form the candidate set.

\noindent\textbf{Instruction Divergence.} 
Since different videos may correspond to similar instructions, we further employ instruction divergence to obtain high-quality data.
Concretely, we randomly select five samples $Q_i=\{Q_i^1,Q_i^2,...,Q_i^5\}$ from the instruction pool of the candidate video set. The instruction distinctiveness $d(V_i,V_j)$ between videos is defined by
\begin{equation}
    \begin{aligned}
        d(V_i,V_j) &= 1-cos(f^{q}_i,f^{q}_j), \\
        \text{where} \:\:\: f^{q}_i &= \text{AvgPool}(F_i^q),
    \end{aligned}
\end{equation}
where $F_i^q$ is the text features of $Q_i$ extracted by the VL embedding models, and $f^{q}_i$ is the averaged representation.
Then we retain the top eight negative video samples for each video, which are visually similar but textually different.
Finally, we concatenate each video with its similar negative videos to form a synthesized long one.

%% file: sec/4_experiment.tex
\section{Experiment}

\subsection{Implementation Details}
\label{sec:id}

In our experiments, we validate our OneClip-RAG with two popular embedding models, which are CLIP-ViTB/32~\cite{RadfordKHRGASAM21} and SigLIP-B/16~\cite{ZhaiM0B23}.
As described in Sec.~\ref{sec:ift}, we first train the embedding model in a coarse-grained manner for $5$ epochs on the QaEgo4D~\cite{BarmannW22} and NeXT-QA~\cite{XiaoSYC21} training sets, and then perform a granular training for $5$ epochs on our constructed SynLongVideo.
We use AdamW~\cite{KingmaB14} to optimize the model with a learning rate of $1e^{-7}$.
We experiment with three popular MLLMs: LLaVA-Video~\cite{ZhangWLLMLL25}, Qwen2.5-VL~\cite{Qwen2.5-VL} and Qwen3-VL~\cite{Qwen3-VL}.
These MLLMs are kept frozen during experiments, of which settings are default.

\subsection{Benchmarks and Metrics}

To validate OneClip-RAG, we conduct extensive experiments on four benchmarks of long video understanding, including MLVU~\cite{abs-2406-04264}, LongVideoBench~\cite{abs-2407-15754}, LVBench~\cite{abs-2406-08035} and Video-MME~\cite{FuDLLRZWZSZCLLZ25}.
MLVU includes videos ranging from 3 minutes to 2 hours. LongVideoBench contains videos up to an hour long. 
The average video duration for LVBench is approximately 68.4 minutes.
Video-MME covers videos of diverse genres and durations, including short, medium, and long-form content.
\emph{Accuracy} (\textbf{Acc}) is used as the evaluation metric for multi-choice VideoQA tasks.

\input{tabs/overall_methodv2}

\input{tabs/overall_sota}

\input{tabs/ab_tc}

\input{tabs/ab_td}

\subsection{Quantitative Analysis}

\noindent\textbf{Comparisons with existing RAG methods on different MLLMs.} 
We first compare our OneClip-RAG with representative video-RAG methods across three MLLMs, with results presented in Tab.~\ref{tb:overall_method}. From Tab.~\ref{tb:overall_method}, we can observe that although directly selecting Top-k keyframes is simple, it is in-fact a strong baseline. But its advantages mainly lie in the local understanding tasks like LVBench, which require precise reasoning over detailed visual information within extended temporal contexts.
In comparison, advanced keyframe RAG methods often exhibit more balanced performance across tasks. For instance, AKS and Q-Frame can obtain decent results on both LVBench and MLVU, mainly due to their adaptive sampling designs.
Compared with these keyframe selection methods, our OneClip-RAG achieve best performance across most benchmarks with average improvements of 11.9\% and 11.4\% on Qwen2.5-VL using CLIP and SigLIP, respectively. 
The consistent improvements across different types of MLLMs demonstrates the superiority and strong generalization of \ours as a plug-and-play component.
Moreover, we can see that \ours is applicable to both CLIP and SigCLIP, showing its generalization on VL embedding models. Overall, these results well confirm the effectiveness of our OneClip-RAG in improving long video understanding of MLLMs.

\noindent\textbf{Comparison with SOTA Video-MLLMs.} 
We further compare \ours with existing SOTA Video-MLLMs on four benchmarks in Tab.~\ref{tb:overall_sota}. 
As shown in Tab.~\ref{tb:overall_sota}, when employing the uniform sampling strategy, short Video-MLLMs, \emph{e.g.,} Keye-VL and GLM-4.1V, achieve superior performance on Video-MME, which primarily requires global understanding capabilities.
However, this straightforward solution proves insufficient for LongVideoBench or LVBench that necessitates fine-grained detail reasoning over extended video sequences.
In contrast, \ours enables Qwen3-VL to achieve SOTA performance across most benchmarks, outperforming both visual compression methods like Video-XL2 and other long-video MLLMs.
This finding can be attributed to \ours's ability to provide more coherent visual context through effective video clip retrieval.
Overall, these results well confirm the advantages of \ours in improving long video understanding capacities of MLLMs.

\begin{figure}[t]
  \centering
   \includegraphics[width=\linewidth]{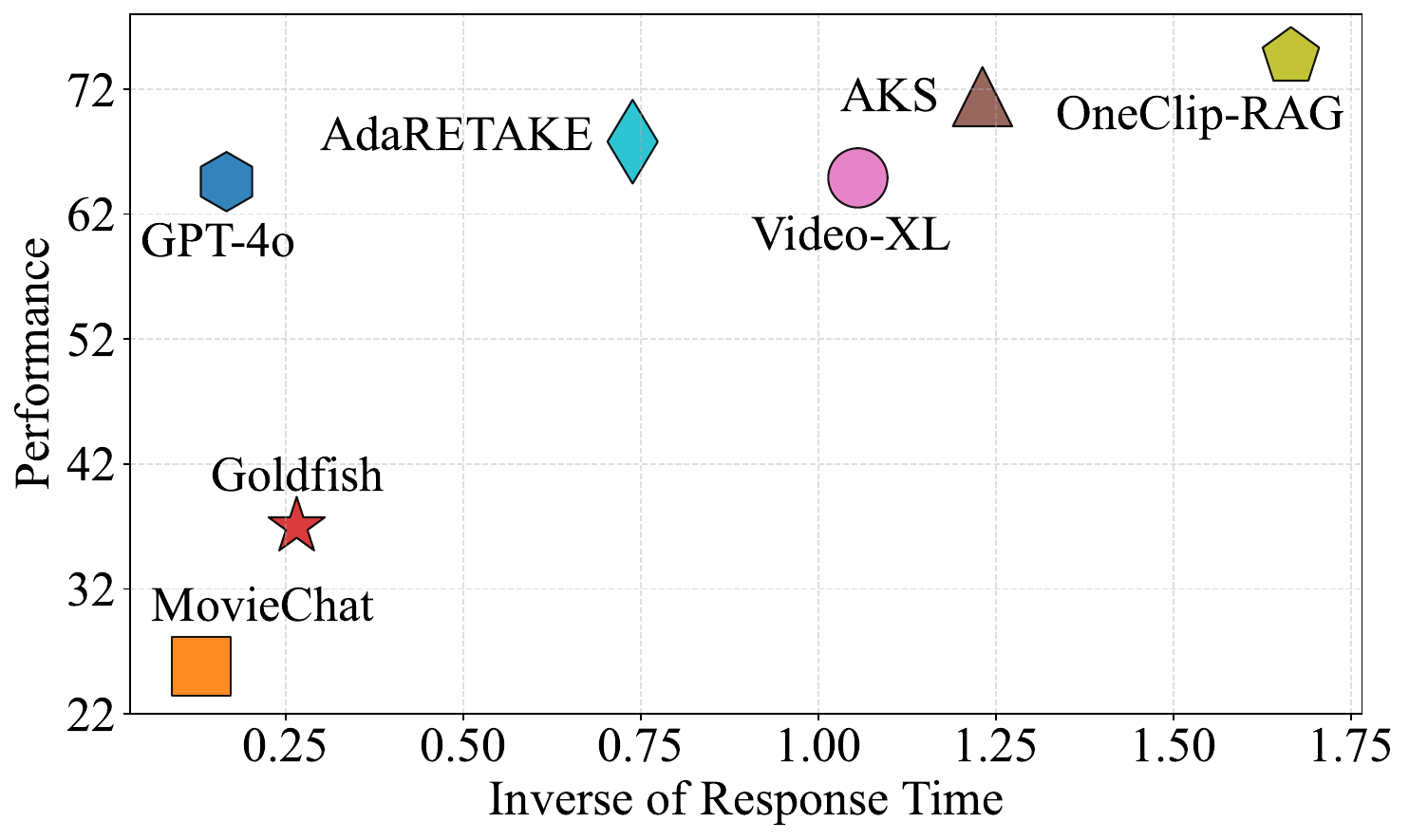}
   \caption{Efficiency and performance comparison between \ours and other SOTA Video-MLLMs on MLVU. OneClip achieves superior performance with greater efficiency.
   }
   \label{fig:efficiency}
\end{figure}

\noindent\textbf{Efficiency of \ours.} 
We further report the time-cost and performance gains of OneClip-RAG in Fig.~\ref{fig:efficiency} and Tab.~\ref{tb:ab_tc}. From Fig.~\ref{fig:efficiency}, we can first observe that the long video-MLLMs, \emph{e.g.}, GPT-4o and Video-XL, typically require several minutes to process videos from MLVU, of which length is about 15 minutes. 
While frame-level VideoRAG methods such as AKS demonstrate much faster inference, \ours demonstrates superiority in both performance and response time, achieving a +2.8 performance gain and a 1.35x speedup over AKS. 
Besides, compared with clip-caption based methods, \emph{i.e.}, Goldfish, our advantages are also very obvious in both performance and efficiency. Due to the additional captioning of uniformly chunked video clips, Goldfish requires several minutes to build the video caption base. In stark contrast, our direct cross-modal retrieval and knowledge augmentation are much more efficiency. 
In Tab.~\ref{tb:ab_tc}, we also report the detailed time costs of all steps in our OneClip-RAG. We can find that the most time-consuming step is the loading of the complete video to memory (20\emph{s}). The actual cost of OneClip-RAG is very cheap since it requires no redundant operations (0.67\emph{s}). Overall, these results well validate the effectiveness and efficiency of our OneClip-RAG towards the long video understanding.

\begin{figure*}[t]
  \centering
   \includegraphics[width=\linewidth]{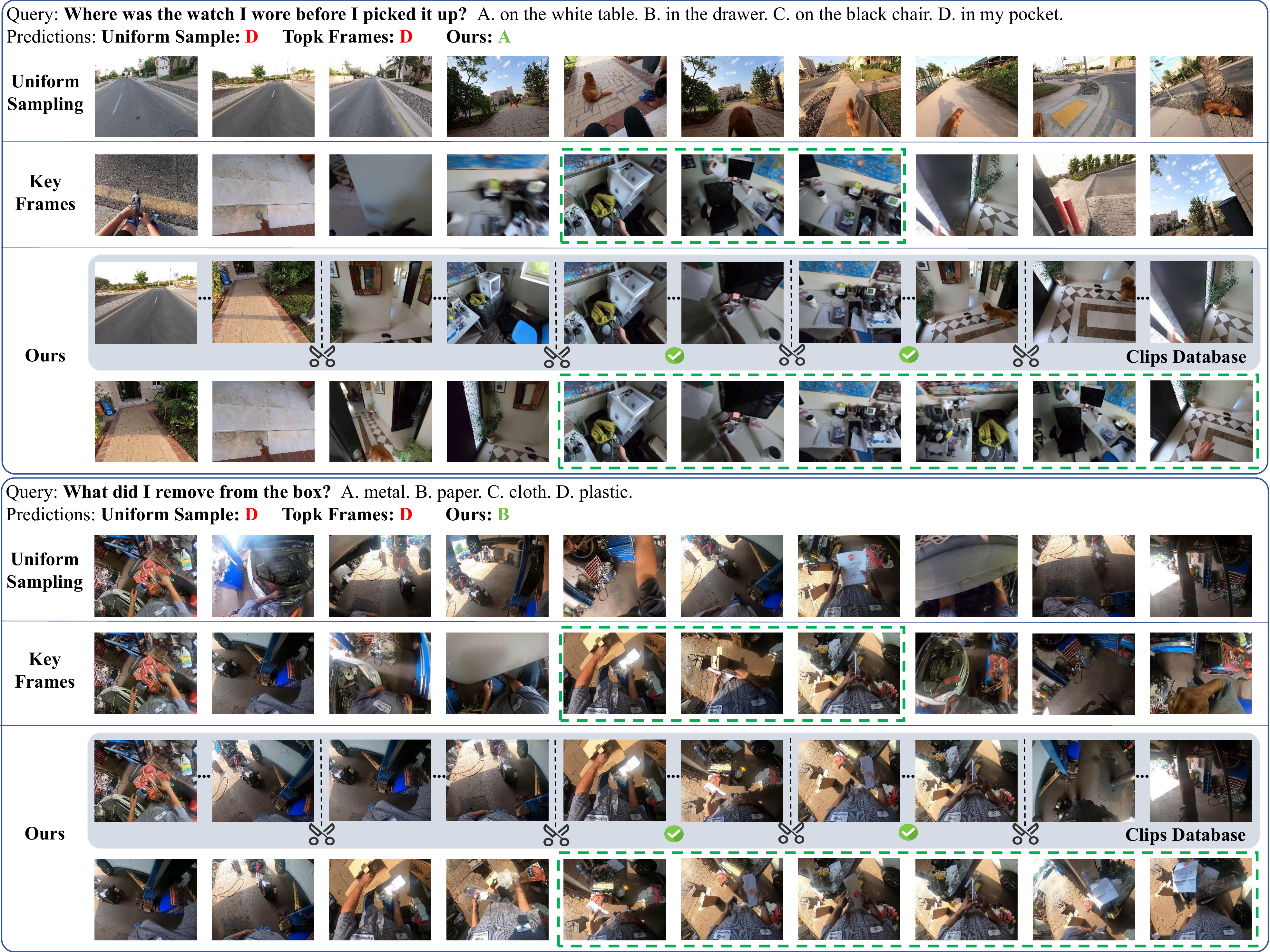}

   \caption{Visualized comparisons between our OneClip-RAG and other Video-RAG methods. The green letters are ground-truth answers, and the green dotted boxes indicate the frames of the long video that is related to the user's instruction.
   }
   \label{fig:visual}

\end{figure*}

\noindent\textbf{Ablation Study.}
We also ablate the key designs and settings of our OneClip-RAG in Tab.~\ref{tb:ab_td}. The results of the first block show the comparison between our use of query-guided clips and the alternative RAG settings.
\emph{Baseline} means uniformly sampling, and \emph{Uniform Clips} denotes the uniform chunking of videos while \emph{Scene Clips} denotes the scene-based ones, \emph{i.e.}, SceneTiling~\cite{abs-2409-01071}.
We can observe that uniform sampling is inferior to others due to the loss of key information. 
Compared with it, \emph{Key Frames} can obtain more relevant visual information for MLLMs, thus showing obvious gains on all benchmarks and metrics. However, the last three clip-based settings can further improve the performance on LongVideoBench, and our query-guided strategy is the best among all comparisons. These results well confirm the merits of clip-RAG as well as the designs of our OneClip-RAG.

In the second block of Tab.~\ref{tb:ab_td}, we examine the effects of our progressive tuning on our SynLongVideo data. We can first observe that the default embedding model (\emph{w/o Training}), \emph{i.e.}, CLIP~\cite{RadfordKHRGASAM21}, lags behind the other two training-based settings. 
Besides, we can see that the direct tuning on short-video instruction data, \emph{i.e.}, \emph{coarse} (Eq.~\ref{eq:cg}), can well improve its question-following capability for better performance. Meanwhile, our full training regime, \emph{i.e.}, \emph{coarse-to-fine}, can further improve the embedding model with fine-grained instruction-clip learning, \emph{i.e.}, Eq.~\ref{eq:fine_loss}, especially for LongVideoBench requiring precise evidence localization.
In the last block of Tab.~\ref{tb:ab_td}, we examine the number of retrieved clips.  
We can observe that \ours can well address the single-detail and multi-detail tasks in MLVU that require detailed information, whereas the holistic task is more suitable for uniform sampling.
Overall, these results further confirm the designs of OneClip-RAG.

\subsection{Qualitative Analysis}

In Fig.~\ref{fig:visual}, we visualize the results of \ours compared with different sampling and retrieval strategies for different VL examples.
As observed, in the context of long video understanding, uniform sampling method typically employed by common MLLMs overlooks key visual information critical to instruction. 
Employing the frame retrieval strategy helps to alleviate this issue. However, this approach yields incoherent video content that include potential semantic conflicts, creating confusion for MLLMs.
However, \ours effectively address this problem by retrieving temporally coherent clips, thereby enabling the video MLLM to make correct precitions.

%% file: tabs/overall_methodv2.tex
\begin{table*}[t]
\centering
\caption{Results of video-MLLMs with \ours on four long VideoQA benchmarks. The best and second-best results are shown in \textbf{bold} and \underline{underlined} respectively.}
\label{tb:overall_method}
\resizebox{\textwidth}{!}{

\renewcommand{\arraystretch}{0.8}
\begin{tabular}{c c c c c c c c c c}
\toprule
\multirow{2}{*}{\textbf{Method}}
& \multirow{2}{*}{\shortstack{\textbf{Embedding}\\\textbf{Model}}}
& \multirow{2}{*}{\textbf{Frames}}
& \multicolumn{2}{c}{\textbf{LongVideoBench}}
& \multicolumn{2}{c}{\textbf{Video-MME}}
& \multirow{2}{*}{\textbf{LVBench}}
& \multirow{2}{*}{\textbf{MLVU}}
& \multirow{2}{*}{\textbf{Avg}}\\
\cmidrule(lr){4-5}
\cmidrule(lr){6-7}
& & & \textbf{Long} & \textbf{Overall} &
 \textbf{Long} & \textbf{Overall} & & & 
\\
\midrule

\multicolumn{1}{l}{\textbf{LLaVA-Video-7B}} & - & 64 
& 49.6 & 58.9 
& 53.2 & 64.2 
& 41.9 & 69.5
& -
\\
\midrule

\textit{Top-k} & CLIP & 64 
& 54.6\,\textcolor{black!45}{(+5.0)} & 59.5\,\textcolor{black!45}{(+0.6)}
& 53.2\,\textcolor{black!45}{(+0.0)} & 64.4\,\textcolor{black!45}{(+0.2)}
& 46.9\,\textcolor{black!45}{(+5.0)}
& 71.2\,\textcolor{black!45}{(+1.7)}
& +3.9\%
\\

\textit{FRAG}~\cite{abs-2504-17447} & MLLM (7B) & 64 
& - & 60.6\,\textcolor{black!45}{(+1.7)}
& - & 63.7\,\textcolor{black!45}{(-0.5)}
& -
& 69.2\,\textcolor{black!45}{(-0.3)}
& +0.6\%
\\

\textit{BOLT}~\citep{00010XG25} & CLIP & 64 
& - & 62.2\,\textcolor{black!45}{(+3.3)}
& - & 64.6\,\textcolor{black!45}{(+0.4)}
& -
& 70.3\,\textcolor{black!45}{(+0.8)}
& +2.5\%
\\

\textit{E-VRAG}~\citep{abs-2508-01546} & VLM (2B) & 64 
& - & \underline{63.1}\,\textcolor{black!45}{(+4.2)}
& - & \textbf{65.4}\,\textcolor{black!45}{(+1.2)}
& -
& 70.2\,\textcolor{black!45}{(+0.7)}
& +3.3\%
\\

\textit{AKS}~\citep{abs-2502-21271} & BLIP & 64 
& 54.7\,\textcolor{black!45}{(+5.1)} & 62.7\,\textcolor{black!45}{(+3.8)}
& \textbf{55.0}\,\textcolor{black!45}{(+1.8)} & \underline{65.3}\,\textcolor{black!45}{(+1.1)}
& 47.6\,\textcolor{black!45}{(+5.7)}
& 71.8\,\textcolor{black!45}{(+2.3)}
& +6.3\%
\\

\textit{Q-Frame$^{\dagger}_{}$}~\cite{abs-2506-22139} & LongCLIP & 64 
& 56.9\,\textcolor{black!45}{(+7.3)} & 61.5\,\textcolor{black!45}{(+2.6)}
& 53.9\,\textcolor{black!45}{(+0.7)} & 64.7\,\textcolor{black!45}{(+0.5)}
& 47.1\,\textcolor{black!45}{(+5.2)}
& 72.4\,\textcolor{black!45}{(+2.9)}
& +5.4\%
\\

\rowcolor{blue!10} \textit{OneClip-RAG} & CLIP & 64 
& \underline{58.0}\,\textcolor{black!45}{(+8.4)} & 62.8\,\textcolor{black!45}{(+3.9)}
& \underline{54.0}\,\textcolor{black!45}{(+0.8)} & 65.2\,\textcolor{black!45}{(+1.0)}
& \underline{49.1}\,\textcolor{black!45}{(+7.2)}
& \textbf{74.6}\,\textcolor{black!45}{(+5.1)}
& \underline{+8.2\%}
\\
\rowcolor{blue!10} \textit{OneClip-RAG} & SigLIP & 64
& \textbf{58.2}\,\textcolor{black!45}{(+8.6)} & \textbf{63.3}\,\textcolor{black!45}{(+4.4)}
& 53.8\,\textcolor{black!45}{(+0.6)} & 64.9\,\textcolor{black!45}{(+0.7)}
& \textbf{50.5}\,\textcolor{black!45}{(+8.6)}
& \underline{73.6}\,\textcolor{black!45}{(+4.1)}
& \textbf{+8.7\%}
\\
\midrule

\multicolumn{1}{l}{\textbf{Qwen2.5-VL-7B}} & - & 64
& 50.7 & 60.1 
& 52.9 & 63.7 
& 39.3 & 65.5
& -
\\
\midrule

\textit{Top-k} & CLIP & 64 
& 56.4\,\textcolor{black!45}{(+5.7)} & 63.5\,\textcolor{black!45}{(+3.4)}
& \underline{54.8}\,\textcolor{black!45}{(+1.9)} & \underline{65.0}\,\textcolor{black!45}{(+1.3)}
& 46.7\,\textcolor{black!45}{(+7.4)}
& 70.7\,\textcolor{black!45}{(+5.2)}
& +8.6\%
\\

\textit{AKS$^\dagger$}~\cite{abs-2502-21271} & BLIP & 64 
& 56.6\,\textcolor{black!45}{(+5.9)} & 63.8\,\textcolor{black!45}{(+3.7)}
& 54.4\,\textcolor{black!45}{(+1.5)} & 64.6\,\textcolor{black!45}{(+0.9)}
& 46.4\,\textcolor{black!45}{(+7.1)}
& 69.6\,\textcolor{black!45}{(+4.1)}
& +8.0\%
\\

\textit{Q-Frame$^{\dagger}_{}$}~\citep{abs-2506-22139} & LongCLIP & 64 
& \underline{58.4}\,\textcolor{black!45}{(+7.7)} & \textbf{64.8}\,\textcolor{black!45}{(+4.7)}
& 54.0\,\textcolor{black!45}{(+1.1)} & 64.5\,\textcolor{black!45}{(+0.8)}
& 46.5\,\textcolor{black!45}{(+7.2)}
& 72.8\,\textcolor{black!45}{(+7.3)}
& +9.6\%
\\

\rowcolor{blue!10} \textit{OneClip-RAG} & CLIP & 64 
& \textbf{58.9}\,\textcolor{black!45}{(+8.2)} & \textbf{64.8}\,\textcolor{black!45}{(+4.7)}
& \textbf{56.3}\,\textcolor{black!45}{(+3.4)} & \textbf{65.3}\,\textcolor{black!45}{(+1.6)}
& \textbf{49.3}\,\textcolor{black!45}{(+10)}
& \underline{73.2}\,\textcolor{black!45}{(+7.7)}
& \textbf{+11.9\%}
\\

\rowcolor{blue!10} \textit{OneClip-RAG} & SigLIP & 64 
& 56.4\,\textcolor{black!45}{(+5.7)} & \underline{64.3}\,\textcolor{black!45}{(+4.2)}
& 54.7\,\textcolor{black!45}{(+1.8)} & \underline{65.0}\,\textcolor{black!45}{(+1.3)}
& \underline{48.1}\,\textcolor{black!45}{(+8.8)}
& \textbf{74.7}\,\textcolor{black!45}{(+9.2)}
& \underline{+11.4\%}
\\
\midrule

\multicolumn{1}{l}{\textbf{Qwen3-VL-8B}} & - & 64 
& 50.7 & 62.2 
& 56.9 & 67.6 
& 43.6 & 71.0
& -
\\
\midrule

\textit{Top-k} & CLIP & 64 
& 57.6\,\textcolor{black!45}{(+6.9)} & 64.1\,\textcolor{black!45}{(+1.9)}
& 57.3\,\textcolor{black!45}{(+0.4)} & 67.7\,\textcolor{black!45}{(+0.1)}
& 49.2\,\textcolor{black!45}{(+5.6)}
& 73.2\,\textcolor{black!45}{(+2.2)}
& +4.8\%
\\

\textit{AKS$^\dagger$}~\citep{abs-2502-21271} & BLIP & 64 
& 56.9\,\textcolor{black!45}{(+8.0)} & 65.2\,\textcolor{black!45}{(+2.4)}
& \textbf{59.0}\,\textcolor{black!45}{(+2.1)} & 68.6\,\textcolor{black!45}{(+1.0)}
& 49.0\,\textcolor{black!45}{(+5.6)}
& 74.2\,\textcolor{black!45}{(+3.2)}
& +5.8\%
\\

\textit{Q-Frame$^{\dagger}_{}$}~\cite{abs-2506-22139} & LongCLIP & 64 
& \underline{58.3}\,\textcolor{black!45}{(+7.6)} & 65.0\,\textcolor{black!45}{(+2.8)}
& 57.2\,\textcolor{black!45}{(+0.3)} & 67.9\,\textcolor{black!45}{(+0.3)}
& 50.2\,\textcolor{black!45}{(+6.6)}
& 74.7\,\textcolor{black!45}{(+3.7)}
& +6.3\%
\\

\rowcolor{blue!10} \textit{OneClip-RAG} & CLIP & 64 
& \textbf{58.5}\,\textcolor{black!45}{(+7.8)} & \textbf{66.3}\,\textcolor{black!45}{(+4.1)}
& 57.8\,\textcolor{black!45}{(+0.9)} & \underline{68.9}\,\textcolor{black!45}{(+1.3)}
& \textbf{54.8}\,\textcolor{black!45}{(+11.2)}
& \textbf{77.1}\,\textcolor{black!45}{(+6.1)}
& \textbf{+10.7\%}
\\

\rowcolor{blue!10} \textit{OneClip-RAG} & SigLIP & 64 
& \underline{58.3}\,\textcolor{black!45}{(+7.6)} & \underline{65.7}\,\textcolor{black!45}{(+3.5)}
& \underline{58.3}\,\textcolor{black!45}{(+1.4)} & \textbf{69.2}\,\textcolor{black!45}{(+1.6)}
& \underline{54.0}\,\textcolor{black!45}{(+10.4)}
& \underline{76.8}\,\textcolor{black!45}{(+5.8)}
& \underline{+10.0\%}
\\

\bottomrule
\end{tabular}

}

\end{table*}

%% file: tabs/overall_sota.tex
\begin{table*}[t]
\centering
\caption{Comparison of SOTA Video-MLLMs and LLaVA-Video with \ours on four long VideoQA benchmarks.}
\label{tb:overall_sota}
\resizebox{\textwidth}{!}{
\begin{tabular}{cccccccccc}
\toprule
\multirow{2}{*}{\textbf{Method}}       & \multirow{2}{*}{\textbf{LLM}} & \multirow{2}{*}{\textbf{Frames}} & \multirow{2}{*}{\textbf{LongVideoBench}} & \multicolumn{4}{c}{\textbf{Video-MME}}                                                                            & \multirow{2}{*}{\textbf{LVBench}} & \multirow{2}{*}{\textbf{MLVU}} \\ \cmidrule(lr){5-8}
                                       &                               &                                  &                                          & \textbf{Short}             & \textbf{Medium}            & \textbf{Long}              & \textbf{Overall}           &                                   &                                \\ \midrule
\rowcolor{gray!20} GPT-5               & -                             & -                                & 72.6                                     &                            &                            & -                          & 81.8                       & -                                 & 77.3                           \\
\rowcolor{gray!20} GPT-4o              & -                             & -                                & 66.7                                     & 80.0                       & 70.3                       & 65.3                       & 71.9                       & -                                 & 64.6                           \\
\rowcolor{gray!20} Gemini-1.5-Pro      & -                             & -                                & 64.0                                     & 81.7                       & 74.3                       & 67.4                       & 75.0                       & 33.1                              & -                              \\ \midrule
Video-XL2~\cite{abs-2506-19225}        & 8B                            & 1fps                             & 61.0                                     & -                          & -                          & -                          & 66.6                       & 48.4                              & 74.8                           \\
Keye-VL~\cite{team2025kwai}            & 8B                            & 0.5fps                           & \underline{62.8}                               & -                          & -                          & -                          & 67.7                       & -                                 & -                              \\
mPLUG-Owl3~\cite{Ye0LH0000025}         & 8B                            & 128                              & 59.7                                     & 70.0                       & 57.7                       & 50.1                       & 59.3                       & 43.5                              & 70.0                           \\
Apollo~\cite{zohar2025apollo}          & 7B                            & 2fps                             & 58.5                                     & -                          & -                          & -                          & 61.3                       & -                                 & 68.7                           \\
VideoNSA~\cite{song2025videonsa}       & 7B                            & 2048                             & 60.0                                     & -                          & -                          & -                          & -                          & -                                 & -                              \\
LongVU~\cite{abs-2410-17434}           & 7B                            & 1fps                             & -                                        & -                          & -                          & \textbf{59.5}              & 60.6                       & -                                 & 65.4                           \\
VideoLLaMA3~\cite{zhang2025videollama} & 7B                            & 1fps                             & 59.8                                     & \textbf{80.1}              & 63.7                       & 54.9                       & 66.2                       & 45.3                              & 73.0                           \\
NVILA~\cite{LiuZSZLYXCGLLTF25}         & 7B                            & 1024                             & 57.7                                     & 75.7                       & 62.2                       & 54.8                       & 64.2                       & -                                 & 70.1                           \\
GLM-4.1V~\cite{hong2025glm}            & 9B                            & -                                & \textbf{65.7}                            & -                          & -                          & -                          & \underline{68.2}                 & 44.0                              & 71.5                           \\
AdaRETAKE~\cite{WangSZWCN25}           & 7B                            & 2fps                             & 62.6                                     & -                          & -                          & \underline{58.3}                 & 67.7                       & \underline{51.2}                        & \underline{75.0}                  \\
ByteVideoLLM~\cite{abs-2412-09530}     & 14B                           & 256                              & -                                        & 74.4                       & 62.9                       & 56.4                       & 64.6                       & -                                 & 70.1                           \\ \midrule
\textcolor{black!50}{Qwen3-VL}         & \textcolor{black!50}{8B}      & \textcolor{black!50}{64}         & \textcolor{black!50}{62.2}               & \textcolor{black!50}{\underline{79.4}} & \textcolor{black!50}{\underline{66.6}} & \textcolor{black!50}{56.9} & \textcolor{black!50}{67.6} & \textcolor{black!50}{43.6}        & \textcolor{black!50}{71.0}     \\
\textbf{+\ours}                        & 8B                            & 64                               & \textbf{65.7}                            & \underline{79.4}                 & \textbf{69.9}              & \underline{58.3}                 & \textbf{69.2}              & \textbf{54.0}                     & \textbf{76.8}                  \\ \bottomrule
\end{tabular}

}

\end{table*}

%% file: tabs/ab_tc.tex
\begin{table*}[t]
\centering
\caption{Detailed computation costs (seconds) for LLaVA-Video with \ours on MLVU and long videos from Video-MME.}
\label{tb:ab_tc}
\resizebox{\textwidth}{!}{
\renewcommand{\arraystretch}{0.9}
\begin{tabular}{c|c|c|cccc|c}
\toprule
Module        & Average Duration   & -             & \multicolumn{4}{c|}{OneClip}                                                  & MLLM      \\ \midrule
Stage          &-  & Video Loading & Feature Extraction & Similarity Calculation & Video Chunking & Clip Retrieval & Inference \\ \midrule
MLVU        &   930$s$  & 20.327        & 1.074             & \textbf{0.001}         & \textbf{0.643}    & \textbf{0.021}      & 2.591     \\ \midrule
Video-MME(Long) &  2466$s$ & 59.810       & 2.074             & \textbf{0.002}         & \textbf{1.737}    & \textbf{0.053}      & 2.997          \\ \bottomrule
\end{tabular}

}
\end{table*}

%% file: tabs/ab_td.tex
\begin{table}[t]
\caption{Ablation studies of key designs of OneClip-RAG. The used settings are marked with $\ddag$ indicate our chosen settings. $^*$We use SceneTiling to obtain scene-based clips. \emph{LVB} denotes the LongVideoBench.}
\label{tb:ab_td}

\setlength{\tabcolsep}{0.6mm}
\renewcommand{\arraystretch}{0.8}

\begin{tabular}{c|cccc|c}
\toprule
\multirow{2}{*}{\textbf{Choices}} & \multicolumn{4}{c|}{\textbf{MLVU}}              & \textbf{LVB} \\ \cmidrule(l){2-6} 
                                  & Single & Multi & Holistic & M-avg & Val          \\ \midrule
\multicolumn{6}{c}{\textbf{Retrieval Strategy}}                                                    \\ \midrule
Baseline                          & 71.7          & 52.0         & 85.3     & 71.0  & 62.2         \\
Key Frames                        & 76.8          & 56.4         & 81.9     & 73.5  & 63.2         \\
Uniform Clips                     & 76.1          & 52.5         & 82.9     & 72.5  & 63.4         \\
Scene Clips$^*$                   & 76.2          & 55.3         & 82.7     & 73.1  & 64.0         \\
Query-Guided Clips$^{\ddag}$      & 77.8          & 68.4         & 84.2     & 77.1  & 66.3         \\ \midrule
\multicolumn{6}{c}{\textbf{Training Strategy}}                                                     \\ \midrule
w/o Training                      & 76.4          & 57.6         & 82.7     & 73.7  & 63.5         \\
Coarse                            & 78.8          & 54.0         & 83.4     & 74.5  & 63.9         \\
Coarse-to-fine$^{\ddag}$          & 77.8          & 68.4         & 84.2     & 77.1  & 66.3         \\ \midrule
\multicolumn{6}{c}{\textbf{Num. of Clips}}                                                         \\ \midrule
4                                 & 73.8          & 47.3         & 77.5     & 69.0  & 62.4         \\
8                                 & 77.4          & 54.0         & 79.7     & 72.9  & 63.4         \\
16                                & 78.1          & 61.9         & 82.9     & 75.7  & 65.4         \\ 
32$^{\ddag}$                      & 77.8          & 68.4         & 84.2     & 77.1  & 66.3         \\ \bottomrule
\end{tabular}%
\end{table}

%% file: sec/5_conclusion.tex
\section{Conclusion}
\label{sec:con}

In this paper, we present OneClip-RAG, a novel and efficient paradigm, for long video understanding of MLLMs. 
The principle of OneClip-RAG is to select instruction-related and continual video clips for the augmented long video understanding. To validated OneClip-RAG, we plug it into five recent MLLMs and conduct extensive experiments on four challenging long video benchmarks. The experimental results not only validate its effecitiveness in improving MLLMs' performance, but also show its great efficiency in handling long video understanding.

\section{Acknowledgments}

This work is supported by the National Key Research and Development Program of China (No. 2025YFE0113500), the National Science Fund for Distinguished Young Scholars (No. 62525605), the National Natural Science Foundation of China (No. U25B2066, No. U22B2051, No.62572407) , Fujian Province Special Science and Technology Program (No. 2025H0041).

\section*{Declarations}

\begin{itemize}
    \item Funding: This work is supported by the National Key Research and Development Program of China (No. 2025YFE0113500), the National Science Fund for Distinguished Young Scholars (No. 62525605), the National Natural Science Foundation of China (No. U25B2066, No. U22B2051, No.62572407) , Fujian Province Special Science and Technology Program (No. 2025H0041).  
    \item Conflict of interest/Competing interests: The authors have no relevant financial or non- financial interests to disclose.
    \item Ethics approval and consent to participate: The authors have no relevant ethics approval to disclose.
    \item Consent for participate: All authors agreed to participate in this work and made clear contributions.
    \item Consent for publication: All authors agreed with the content and that all gave explicit consent to submit and that they obtained consent from the responsible authorities at the institute/organization where the work has been carried out.
    \item Data availability: The datasets used during the current study are available in these repositories:

    QaEgo4D~\cite{BarmannW22} \url{https://github.com/lbaermann/qaego4d},
    
    NeXT-QA~\cite{XiaoSYC21} \url{https://huggingface.co/datasets/lmms-lab/NExTQA},

    MLVU~\cite{abs-2406-04264} \url{https://huggingface.co/datasets/MLVU/MVLU},

    LongVideoBench~\cite{abs-2407-15754} \url{https://huggingface.co/datasets/longvideobench/LongVideoBench}, 
    
    LVBench~\cite{abs-2406-08035} \url{https://huggingface.co/datasets/zai-org/LVBench},
    
    Video-MME~\cite{FuDLLRZWZSZCLLZ25} \url{https://huggingface.co/datasets/lmms-lab/Video-MME}.
    
    \item Code availability: Our code is publicly released at: \url{https://github.com/Tao-Chen-xmu/OneClip-RAG}.
    \item Author contribution: Yiyi Zhou determined the research objectives, proposed the primary research direction, provided guidance on methodology, and led the revision of the manuscript. Tao Chen designed the specific methodology. Shaobo Ju, Qiong Wu, Chenxin Fang, and Kun Zhang participated in discussions and assisted in the implementation, experiments, and data analysis. Tao Chen wrote the first draft of the manuscript. All authors read, commented on previous versions, and approved the final manuscript. More detailed contributions of each author are listed below: Conceptualization: Yiyi Zhou; Methodology: Tao Chen; Investigation and Implementation: Shaobo Ju, Qiong Wu, Chenxin Fang, and Kun Zhang;  Writing - original draft preparation: Tao Chen; Writing - review and editing: Yiyi Zhou, Hui Li, Jun Peng, and Rongrong Ji; Supervision: Rongrong Ji.
\end{itemize}